\documentclass{article} % For LaTeX2e
\usepackage{iclr2020_conference,times}

\usepackage{amsmath,amsfonts,bm}

\def\eqref#1{equation~\ref{#1}}
\def\1{\bm{1}}

\def\va{{\bm{a}}}

\def\vv{{\bm{v}}}

\def\mW{{\bm{W}}}

\DeclareMathAlphabet{\mathsfit}{\encodingdefault}{\sfdefault}{m}{sl}
\SetMathAlphabet{\mathsfit}{bold}{\encodingdefault}{\sfdefault}{bx}{n}

\def\gN{{\mathcal{N}}}

\newcommand{\R}{\mathbb{R}}

\newcommand{\normlone}{L^1}

\usepackage{url}
\usepackage{graphicx}
\usepackage{subfigure}
\usepackage{threeparttable}
\usepackage{amsmath}
\usepackage{color}
\usepackage{wrapfig}
\usepackage{booktabs}
\usepackage{multirow}
\usepackage{capt-of}
\usepackage[ruled]{algorithm2e}

\title{Fast Neural Network Adaptation via  Parameter Remapping and Architecture Search}

\author{
Jiemin Fang$^{1}$\footnotemark[1]\, \footnotemark[2] , Yuzhu Sun$^{1}$\footnotemark[1]\, \footnotemark[2] , Kangjian Peng$^{2}$\footnotemark[1] , Qian Zhang$^{2}$, Yuan Li$^{2}$, \\
\textbf{ Wenyu Liu$^{1}$, Xinggang Wang$^{1}$\footnotemark[3]} \\
$^1$School of EIC, Huazhong University of Science and Technology $\; ^2$Horizon Robotics\\
\texttt{\{jaminfong, yzsun, liuwy, xgwang\}@hust.edu.cn}\\
\texttt{\{kangjian.peng, qian01.zhang, yuan.li\}@horizon.ai}
}

\definecolor{citegreen}{rgb}{0,0.5,0}
\usepackage[colorlinks=true, linkcolor=red, anchorcolor=blue, citecolor=citegreen]{hyperref}

\iclrfinalcopy % Uncomment for camera-ready version, but NOT for submission.
\begin{document}

\maketitle

\begin{abstract}
    Deep neural networks achieve remarkable performance in many computer vision tasks. Most state-of-the-art~(\emph{SOTA}) semantic segmentation and object detection approaches reuse neural network architectures designed for image classification as the backbone, commonly pre-trained on ImageNet. However, performance gains can be achieved by designing network architectures specifically for detection and segmentation, as shown by recent neural architecture search (NAS) research for detection and segmentation. One major challenge though, is that ImageNet pre-training of the search space representation (a.k.a.~super network) or the searched networks incurs huge computational cost. In this paper, we propose a Fast Neural Network Adaptation (FNA) method, which can adapt both the architecture and parameters of a seed network (e.g. a high performing manually designed backbone) to become a network with different depth, width, or kernels via a Parameter Remapping technique, making it possible to utilize NAS for detection/segmentation tasks a lot more efficiently. In our experiments, we conduct FNA on MobileNetV2 to obtain new networks for both segmentation and detection that clearly out-perform existing networks designed both manually and by NAS. The total computation cost of FNA is significantly less than \emph{SOTA} segmentation/detection NAS approaches: 1737$\times$ less than DPC, 6.8$\times$ less than Auto-DeepLab and 7.4$\times$ less than DetNAS. The code is available at \url{https://github.com/JaminFong/FNA}.
\end{abstract}
\renewcommand{\thefootnote}{\fnsymbol{footnote}}
\footnotetext[1]{Equal contributions.}
\footnotetext[2]{The work is performed during an internship at Horizon Robotics.}
\footnotetext[3]{Corresponding author.}

% --------------------------------Section Partition------------------------------- %
\section{Introduction}
Deep convolutional neural networks have achieved great successes in computer vision tasks such as image classification~\citep{krizhevsky2012imagenet, he2016deep, howard2017mobilenets}, semantic segmentation~\citep{long2015fully, ronneberger2015u, deeplab-v3} and object detection~\citep{DBLP:journals/pami/RenHG017, liu2016ssd, lin2017focal} etc. Image classification has always served as a fundamental task for neural architecture design. It is common to use networks designed and pre-trained on the classification task as the backbone and fine-tune them for segmentation or detection tasks. However, the backbone plays an important role in the performance on these tasks and the difference between these tasks calls for different design principles of the backbones. For example, segmentation tasks require high-resolution features and object detection tasks need to make both localization and classification predictions from each convolutional feature. Such distinctions make neural architectures designed for classification tasks fall short. Some attempts~\citep{DBLP:conf/eccv/LiPYZDS18, DBLP:journals/corr/abs-1908-07919} have been made to tackle this problem.

Handcrafted neural architecture design is inefficient, requires a lot of human expertise, and may not find the best-performing networks. Recently, neural architecture search (NAS) methods~\citep{zoph2017learning, pham2018efficient, liu2017progressive} see a rise in popularity. Some works~\citep{liu2019auto, zhang2019customizable, DBLP:journals/corr/abs-1903-10979} propose to use NAS to design backbone architectures specifically for segmentation or detection tasks. Nevertheless, pre-training remains a inevitable but costly procedure. Though some~\citep{DBLP:journals/corr/abs-1811-08883} recently demonstrates that pre-training is not always necessary for accuracy considerations, training from scratch on the target task still takes more iterations than fine-tuning from a pre-trained model. For NAS methods, the pre-training cost is non-negligible for evaluating the networks in the search space. One-shot search methods~\citep{brock2017smash, Understanding, DBLP:journals/corr/abs-1903-10979} integrate all possible architectures in one super network but pre-training the super network and the searched network still bears huge computation cost.

\begin{figure*}[tbp]
\vspace{-5pt}
    \begin{center}
        \includegraphics[width=1.0\linewidth]{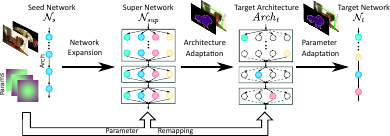}
    \end{center}
    \caption{The framework of our proposed FNA. Firstly, we select an artificially designed network as the seed network $\gN_s$ and expand $\gN_s$ to a super network $\gN_{sup}$ which is the representation of the search space. Then parameters of $\gN_s$ are remapped to $\gN_{sup}$. We utilize the NAS method to start the architecture adaptation with the super network and obtain the target architecture $\mathit{Arch}_t$. Before parameter adaptation, we remap the parameters of $\gN_s$ to $\mathit{Arch}_t$. Finally, we adapt the parameters of $\mathit{Arch}_t$ to get the target network $\gN_t$.}
    \label{fig: framework}
    \vspace{-5pt}
\end{figure*}

As ImageNet~\citep{imagenet} pre-training has been a standard practice for many computer vision tasks, there are lots of models trained on ImageNet available in the community. To take full advantages of these models, we propose a Fast Neural Network Adaptation (FNA) method based on a novel parameter remapping paradigm. Our method can adapt both the architecture and parameters of one network to a new task with negligible cost. Fig.~\ref{fig: framework} shows the whole framework. The adaptation is performed on both the architecture- and parameter-level. We adopt the NAS methods~\citep{zoph2017learning, Real2018Regularized, liu2018darts} to implement the architecture-level adaptation. We select a manually designed network (MobileNetV2~\citep{sandler2018mobilenetv2} in our experiments) as the \verb+seed network+, which is pre-trained on ImageNet. Then, we expand the seed network to a super network which is the representation of the search space in FNA. New parameters in the super network are initialized by mapping those from the seed network using parameter remapping. Thanks to that, the neural architecture search can be performed efficiently on the detection and segmentation tasks. With FNA we obtain a new optimal target architecture for the new task. Similarly, we remap the parameters of the seed network to the target architecture for initialization and fine-tune it on the target task with no need of pre-training on a large-scale dataset.

We demonstrate FNA's effectiveness and efficiency via experiments on both segmentation and detection tasks. We adapt the manually designed network MobileNetV2~\citep{sandler2018mobilenetv2} to segmentation framework DeepLabv3~\citep{deeplab-v3}, detection framework RetinaNet~\citep{lin2017focal} and SSDLite~\citep{liu2016ssd, sandler2018mobilenetv2}. Networks adapted by FNA surpass both manually designed and NAS searched networks in terms of both performance and model MAdds. Compared to NAS methods, FNA costs 1737$\times$ less than DPC~\citep{DBLP:conf/nips/ChenCZPZSAS18}, 6.8$\times$ less than Auto-DeepLab~\citep{liu2019auto} and 7.4$\times$ less than DetNAS~\citep{DBLP:journals/corr/abs-1903-10979}.

% --------------------------------Section Partition------------------------------- %
\section{Related Work}
    \vspace{-5pt}
    \paragraph{Neural Architecture Search}
    With reinforcement learning (RL) and evolutionary algorithm (EA) being applied to NAS methods, many works~\citep{zoph2016neural, zoph2017learning, Real2018Regularized} make great progress in promoting the performances of neural networks. Recently, NAS methods based on one-shot model~\citep{brock2017smash, Understanding} or differentiable representations~\citep{liu2018darts, cai2018proxylessnas, fang2019densely} achieve remarkable results with low search cost. We use the differentiable NAS method to implement architecture adaptation, which adjusts the architecture of the backbone network automatically.

    \vspace{-5pt}
    \paragraph{Backbone Design}
    As deep neural network designing~\citep{simonyan2014very, szegedy2016rethinking, he2016deep} develops, the backbones of segmentation or detection networks evolve accordingly. Some works improve the backbone architectures by modifying existing networks. PeleeNet~\citep{wang2018pelee} proposes a variant of DenseNet~\citep{huang2017densely} for more real-time object detection on mobile devices. DetNet~\citep{DBLP:conf/eccv/LiPYZDS18} applies dilated convolution~\citep{yu2015multi} in the backbone to enlarge the receptive field which helps to detect objects. BiSeNet~\citep{yu2018bisenet} and HRNet~\citep{DBLP:journals/corr/abs-1908-07919} design multiple paths to learn both high- and low- resolution representations for better dense prediction.
    
    \vspace{-5pt}
    \paragraph{Parameter Remapping}
    Net2Net~\citep{chen2015net2net} proposes the function-preserving transformations to remap the parameters of one network to a new deeper or wider network. This remapping mechanism accelerates the training of the new larger network and achieves great performances. Following this manner, EAS~\citep{cai2018efficient} extends the parameter remapping concept to neural architecture search. Moreover, some NAS works~\citep{pham2018efficient, eatnas, elsken2018efficient} apply parameters sharing on child models to accelerate the search process. Our parameter remapping paradigm extends the mapping dimension with the kernel level. Parameters can be also mapped to a shallower or narrower network with our scheme, while Net2Net focuses on mapping parameters to a deeper and wider network. The parameter remapping in our FNA largely decreases the computation cost of the network adaptation by taking full advantages of the ImageNet pre-trained parameters.

% --------------------------------Section Partition------------------------------- %
\section{Method}
As the most commonly used network for designing search spaces in NAS methods~\citep{MnasNet, cai2018proxylessnas, fang2019densely},  MobileNetV2~\citep{sandler2018mobilenetv2} is selected as the seed network to give the details of our method. To adapt the network to segmentation and detection tasks, we adjust the architecture elements on three levels, i.e., convolution kernel size, depth and width of the network. In this section, we first describe the parameter remapping paradigm. Then we explain the whole procedure of the network adaptation.

\subsection{Parameter Remapping}
\label{subsec: params_remap}
    We define \emph{parameter remapping} as one paradigm which aims at mapping the parameters of one seed network to another one. We denote the seed network as $\gN_s$ and the new network as $\gN_n$, whose parameters are denoted as $\mW_s$ and $\mW_n$ respectively. The remapping paradigm is illustrated in the following three aspects. The remapping on the depth-level is firstly carried out and then the remapping on the kernel- and width- level is conducted simultaneously. Moreover, we study different remapping strategies in the experiments (Sec.~\ref{subsec: studyPR}).

\begin{figure*}
    \vspace{-10pt}
    \centering
    \subfigure[Depth level]
    {\begin{minipage}[b]{0.32\textwidth}
        \includegraphics[width=0.9\textwidth]{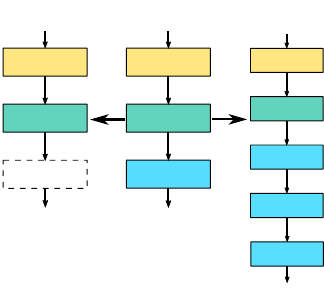}
        \label{fig: depth_mapping}
    \end{minipage}}
    \hfill
    \subfigure[Width level]
    {\begin{minipage}[b]{0.17\textwidth}
        \includegraphics[width=0.9\textwidth]{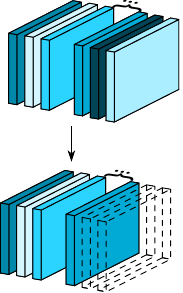}
        \label{fig: channel_mapping}
    \end{minipage}}
    \hfill
    \subfigure[Kernel level]
    {\begin{minipage}[b]{0.25\textwidth}
        \includegraphics[width=0.9\textwidth]{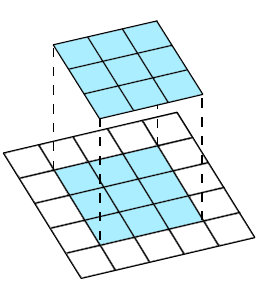}
        \label{fig: kernel_mapping}
    \end{minipage}}
    \vspace{-5pt}
    \caption{Our parameters are remapped on three levels. (a) shows the depth-level remapping. The parameters of existing corresponding layers are mapped from the original network. The parameters of new layers are mapped from the last layer in the original network. (b) shows the width-level remapping. For each channels-diminished dimension, the parameters are copied from the original existing ones. (c) shows the kernel-level remapping. The original parameters are mapped to the central part of the new larger kernel. The values of the other parameters are assigned with 0.}
    \vspace{-5pt}
\end{figure*}

    \paragraph{Remapping on Depth-level}
    We introduce more depth settings in our architecture adaptation process. In other word, we adjust the number of inverted residual blocks (MBConvs)~\citep{sandler2018mobilenetv2} in every stage of the network. We assume that one stage in the seed network $\gN_s$ has $l$ layers. The parameters of each layer can be denoted as $\{\mW_s^{(1)}, \mW_s^{(2)}, \dots, \mW_s^{(l)}\}$. Similarly, we assume that the corresponding stage with $m$ layers in the new network $\gN_n$ has parameters $\{\mW_n^{(1)}, \mW_n^{(2)}, \dots, \mW_n^{(m)}\}$. The remapping process on the depth-level is shown in Fig.~\ref{fig: depth_mapping}. The parameters of layers in $\gN_n$ which also exit in $\gN_s$ are just copied from $\gN_s$. The parameters of new layers are all copied from the last layer in the stage of $\gN_s$. It is formulated as
    \begin{equation}
        \begin{aligned}
            f(i) &= min(i, l),  \\
            W_n^{(i)} &= W_s^{(f(i))},\quad \forall 1 \le i \le m \text{.}
        \end{aligned}
        \label{eq: depth_mapping}
    \end{equation}

    \paragraph{Remapping on Width-level}
    In the MBConv block of MobileNetV2~\citep{sandler2018mobilenetv2} network, the first point-wise convolution expands the low-dimensional features to a high dimension. This practice can be utilized for expanding the width and capacity of one neural network. We allow smaller expansion ratios for architecture adaptation. We denote the parameters of one convolution in $\gN_s$ as $\mW^s \in \R^{p \times q \times h \times w}$ and that in $\gN_n$ as $\mW^n \in \R^{r \times s \times h \times w}$, where $r \le p$ and $s \le q$. As shown in Fig.~\ref{fig: channel_mapping}, on the width-level, we directly map the first $r$ or $s$ channels of parameters in $\gN_s$ to the narrower one in $\gN_n$. It can be formulated as
    \begin{equation}
        \mW^n_{i,j,:,:} = \mW^s_{i,j,:,:}, \quad \forall 1 \le i \le r, 1 \le j \le s\text{.}
        \label{ea: channel_mapping}
    \end{equation}
    
    \paragraph{Remapping on Kernel-level}
    The kernel size is commonly set as $3\times3$ in most artificially-designed networks. To expand the receptive field and capture abundant features in segmentation or detection tasks, we introduce larger kernel size settings in the adaptation process. As Fig.~\ref{fig: kernel_mapping} shows, to expand the $3\times3$ kernel to a larger one, we assign the parameters of the central $3\times3$ region in the large kernel with the values of the original $3\times3$ kernel. The values of the other region surrounding the central part are assigned with 0. We denote the parameters of the original $3\times3$ kernel as $\mW^{3\times3}$ and the larger $k \times k$ kernel as $\mW^{k \times k}$. The remapping process on kernel-level can be formulated as follows,
    \begin{equation}
        \mW^{k \times k}_{:,:,h,w} = \begin{cases}
            \mW^{3 \times 3}_{:,:,h,w} & \text{if} (k-3)/2 < h, w \le (k+3)/2\\
            0 & \text{otherwise}
        \end{cases}\text{,}
        \label{eq: kernel_mapping}
    \end{equation}
    where $h, w$ denote the indices of the spatial dimension. This design principle conforms to the function-preserving concept~\citep{chen2015net2net}, which accelerates and stabilizes the optimization of the new network. 
    
\subsection{Neural Network Adaptation}
We divide our neural network adaptation into three steps. Fig.~\ref{fig: framework} demonstrates the whole adaptation procedure. Firstly, we expand the seed network to a super network which is the representation of the search space in the latter architecture adaptation process. Secondly, we perform the differentiable NAS method to implement network adaptation on the architecture-level and obtain the target architecture $\mathit{Arch}_t$. Finally, we adapt the parameters of the target architecture and obtain the target network $\gN_t$. The aforementioned parameter remapping mechanism is deployed before the two stages, i.e. architecture adaptation and parameter adaptation.

    \vspace{-3pt}
    \paragraph{Network Expansion}
    We expand the seed network to a super network by introducing more options for architecture elements. For every MBConv layer, we allow for more kernel size settings $\{3, 5, 7\}$ and more expansion ratios $\{3, 6\}$. As most differentiable NAS methods~\citep{liu2018darts, cai2018proxylessnas, fbnet} do, we relax every layer as a weighted sum of all candidate operations.
    \begin{equation}
        \overline{o}^{(i)}(x) = \sum_{o \in O} \frac{exp(\alpha^{(i)}_{o})}{\sum_{o' \in O} exp(\alpha^{(i)}_{o'})}o(x)\text{,}
    \end{equation}
    where $O$ denotes the operation set, $\alpha_o^{i}$ denotes the architecture parameter of operation $o$ in the $i$th layer, and $x$ denotes the input tensor. We set more layers in one stage of the super network and add the identity connection to the candidate operation set for depth search. After expanding the seed network to a super network, we remap the parameters of the seed network to the super network based on the paradigm illustrated in Sec.~\ref{subsec: params_remap}. This remapping strategy prevents the huge cost of ImageNet pre-training involved in the search space, i.e. the super network in differentiable NAS.
    
    \vspace{-3pt}
    \paragraph{Architecture Adaptation}
    We start the differentiable NAS process with the expanded super network directly on the target task, e.g., semantic segmentation or object detection. We first fine-tune operation weights of the super network for some epochs on the training dataset. After the weights are sufficiently trained, we start alternating the optimization of operation weights $w$ with $\frac{\partial \mathcal{L}}{\partial w}$ and architecture parameters $\alpha$ with $\frac{\partial \mathcal{L}}{\partial \alpha}$. To accelerate the search process and decouple the parameters of different sub-networks, we only sample one path in each iteration according to the distribution of architecture parameters for operation weight updating. As the search process terminates, we use the architecture parameters $\alpha$ to derive the target architecture.
    
    \vspace{-3pt}
    \paragraph{Parameter Adaptation}
    We obtain the target architecture $\mathit{Arch}_t$ from architecture adaptation. To accommodate the new segmentation or detection tasks, the target architecture becomes  different from that of the seed network $\gN_s$ (which is primitively designed for the image classification task). Unlike conventional training strategy, we discard the cumbersome pre-training process of $\mathit{Arch}_t$ on ImageNet. We remap the parameters of $\gN_s$ to $\mathit{Arch}_t$ utilizing the method described in Sec.~\ref{subsec: params_remap}. Finally, we directly fine-tune $\mathit{Arch}_t$ on the target task and obtain the final target network $\gN_t$.

% --------------------------------Section Partition------------------------------- %
\section{Experiments}

We select the ImageNet pre-trained MobileNetV2 model as the seed network and apply our FNA method on both semantic segmentation and object detection tasks. In this section, we firstly give the implementation details of our experiments; then we report and analyze the network adaptation results; finally, we perform ablation studies to validate the effectiveness of the parameter remapping paradigm and compare different parameter remapping implementations.

\begin{table*}[tbp]
    \vspace{-5pt}
    \centering
    \caption{Semantic segmentation results on Cityscapes. OS: output stride, the spatial resolution ratio of input image to backbone output. The result of DPC in the brackets is our implemented version under the same settings as FNA. The MAdds of the models are computed with the $1024 \times 2048$ input resolution.}
    \label{tab: seg}
    \begin{threeparttable}
    \resizebox{\textwidth}{!}{
        \begin{tabular}{l | c | c | c | c | c | c}
        \hline
        \multicolumn{2}{l|}{\textbf{Method}} & \textbf{OS} &\textbf{iters} & \textbf{Params} & \textbf{MAdds} & \textbf{mIOU(\%)} \\
        \hline
        MobileNetV2~\citep{sandler2018mobilenetv2} & \multirow{3}*{DeepLabv3} &\multirow{3}*{16}& \multirow{3}*{100K} & 2.57M & 24.52B & 75.5 \\
        DPC~\citep{DBLP:conf/nips/ChenCZPZSAS18} & && & 2.51M & 24.69B & 75.4(75.7)\\
        FNA & & && 2.47M & 24.17B & \textbf{76.6}\\
        \hline
        Auto-DeepLab-S~\citep{liu2019auto} & \multirow{3}*{DeepLabv3+} & 8 & 500K & 10.15M & 333.25B& 75.2\\
        FNA & & 16& 100K & 5.71M & 210.11B & 77.2 \\
        FNA & & 8 & 100K & 5.71M & 313.87B & \textbf{78.0} \\
        \hline
    \end{tabular}}
    \end{threeparttable}
    \vspace{-5pt}
\end{table*}

\begin{table*}[tbp]
    \centering
    \caption{Comparisons of computational cost on the semantic segmentation tasks. ArchAdapt: Architecture Adaptation. ParamAdapt: Parameter Adaptation. GHs: GPU Hours. $\ast$ indicates the computational cost calculated under our reproducing settings. $\dagger$ indicates the cost estimated according to the description in the original paper.}
    \label{tab: seg cost}
     \begin{threeparttable}
    \resizebox{0.9\textwidth}{!}{
        \begin{tabular}{l c c c}
        \toprule
        \textbf{Method} & \textbf{Total Cost} & \textbf{ArchAdapt Cost} & \textbf{ParamAdapt Cost} \\
        \midrule
        DPC~\citep{DBLP:conf/nips/ChenCZPZSAS18} & 62.2K GHs & 62.2K GHs & 30.0$^\ast$ GHs\\
        Auto-DeepLab-S~\citep{liu2019auto} & 244.0 GHs & 72.0 GHs & 172.0$^\dagger$ GHs\\
        FNA & 35.8 GHs & 1.4 GHs & 34.4 GHs\\
        \bottomrule
    \end{tabular}
    }
    \end{threeparttable}
    \vspace{-5pt}
\end{table*}

\subsection{Network Adaptation on Semantic Segmentation}

The semantic segmentation experiments are conducted on the Cityscapes~\citep{DBLP:conf/cvpr/CordtsORREBFRS16} dataset. In the architecture adaptation process, we map the seed network to the super network, which is used as the backbone of DeepLabv3~\citep{deeplab-v3}. We randomly sample $20\%$ images from the training set as the validation set for architecture parameters updating. The original validation set is not used in the search process. To optimize the MAdds of the searched network, we define the loss function in search as $\mathcal{L} = \mathcal{L}_{task} + \lambda \log_{\tau}(cost)$. The first term denotes the cross-entropy loss and the second term controls the MAdds of the network. We set $\lambda$ as $9 \times 10^{-3}$ and $\tau$ as $45$. The search process takes $80$ epochs in total. The architecture optimization starts after $30$ epochs. The whole search process is conducted on a single V100 GPU and takes only 1.4 hours in total. 

In the parameter adaptation process, we remap the parameters of original MobileNetV2 to the target architecture obtained in the aforementioned architecture adaptation. The whole parameter adaptation process is conducted on $4$ TITAN-Xp GPUs and takes $100$K iterations, which cost only $8.5$ hours in total.

Our semantic segmentation results are shown in Tab.~\ref{tab: seg}. The FNA network achieves $76.6\%$ mIOU on Cityscapes with the DeepLabv3~\citep{deeplab-v3} framework, $1.1\%$ mIOU better than the manually designed seed Network MobileNetV2~\citep{sandler2018mobilenetv2} with fewer MAdds. Compared with a NAS method DPC~\citep{DBLP:conf/nips/ChenCZPZSAS18} (with MobileNetV2 as the backbone) which searches a multi-scale module for semantic segmentation tasks, FNA gets $0.9\%$ mIOU promotion with $0.52$B fewer MAdds. For fair comparison with Auto-DeepLab~\citep{liu2019auto} which searches the backbone architecture on DeepLabv3 and retrains the searched network on DeepLabv3+~\citep{chen2018encoder}, we adapt the parameters of the target architecture $\mathit{Arch}_t$ to DeepLabv3+ framework. Comparing with Auto-DeepLab-S, FNA achieves far better mIOU with fewer MAdds, Params and training iterations. With the remapping mechanism, FNA only takes 35.8 GPU hours, 1737$\times$ less than DPC and 6.8$\times$ less than Auto-DeepLab.

\begin{table*}[tbp]
    \vspace{-5pt}
    \centering
    \caption{Object detection results on MS-COCO. The MAdds are calculated with $1088 \times 800$ input.}
    \label{tab: det}
    \begin{threeparttable}
    \resizebox{0.9\textwidth}{!}{
        \begin{tabular}{l | c | c | c | c }
        \hline
        \multicolumn{2}{l|}{\textbf{Method}} & \textbf{Params} & \textbf{MAdds} & \textbf{mAP(\%)} \\
        \hline
        ShuffleNetV2-20~\citep{DBLP:journals/corr/abs-1903-10979} & \multirow{4}*{RetinaNet} & 13.19M &    132.76B & 32.1 \\
        MobileNetV2~\citep{sandler2018mobilenetv2} & & 11.49M & 133.05B & 32.8\\
        DetNAS~\citep{DBLP:journals/corr/abs-1903-10979} & & 13.41M & 133.26B & 33.3\\
        FNA & & 11.73M & 133.03B & \textbf{33.9} \\
        \hline
        MobileNetV2~\citep{sandler2018mobilenetv2} &\multirow{3}*{SSDLite}&4.3M &0.8B &22.1 \\
        Mnasnet-92~\citep{MnasNet} & &5.3M &1.0B &22.9 \\
        FNA &&4.6M &0.9B & \textbf{23.3} \\% \textbf{23.0} \\
        \hline
    \end{tabular}}
    \end{threeparttable}
    \vspace{-5pt}
\end{table*}

\begin{table*}[tbp]
    \vspace{-5pt}
    \centering
    \caption{Comparison of computational cost on the object detection tasks. All our experiments on object detection are conducted on TITAN-Xp GPUs.}
    \label{tab: det cost}
    \begin{threeparttable}
    \resizebox{1.0 \textwidth}{!}{
        \begin{tabular}{l| c| c |c| c| c| c}
        \hline
        \multirow{2}*{\textbf{Method}} & \multirow{2}*{\textbf{Total Cost}} & \multicolumn{3}{c|}{\textbf{Super Network}} & \multicolumn{2}{c}{\textbf{Target Network}} \\
        \cline{3-7}
        & & Pre-training &Finetuning &Search &Pre-training & Finetuning \\
        \hline
        DetNAS~\citep{DBLP:journals/corr/abs-1903-10979} &68 GDs &12 GDs &12 GDs &20 GDs &12 GDs &12 GDs \\
        FNA (RetinaNet) &9.2 GDs & - & - & 6 GDs & - &3.2 GDs \\
        FNA (SSDLite) &21.6 GDs & - & - & 6.6 GDs & - &15 GDs \\
        \hline
        \end{tabular}}
    \end{threeparttable}
    \vspace{-5pt}
\end{table*}

\subsection{Network Adaptation on Object Detection}
We further implement our FNA method on object detection tasks. We adapt the MobileNetV2 seed network to two commonly used detection systems, RetinaNet~\citep{lin2017focal} and a lightweight one SSDLite~\citep{liu2016ssd, sandler2018mobilenetv2}. The experiments are conducted on the MS-COCO dataset~\citep{DBLP:conf/eccv/LinMBHPRDZ14}. Our implementation is based on the MMDetection~\citep{chen2019mmdetection} framework. In the search process of architecture adaptation, we randomly sample $50\%$ data from the original \texttt{trainval35k} set as the validation set.

We show the results on the COCO dataset in Tab.~\ref{tab: det}. In the RetinaNet framework, compared with two manually designed networks, ShuffleNetV2-10~\citep{DBLP:journals/corr/abs-1807-11164, DBLP:journals/corr/abs-1903-10979} and MobileNetV2~\citep{sandler2018mobilenetv2}, FNA achieves higher mAP with similar MAdds. Compared with DetNAS~\citep{DBLP:journals/corr/abs-1903-10979} which searches the backbone of detection network, FNA achieves $0.6\%$ higher mAP with $1.64$M fewer Params and $0.2$B fewer MAdds. As shown in Tab.~\ref{tab: det cost}, our total computation cost is only 13.5\% of DetNAS. Moreover, we implement our FNA method on the SSDLite framework. In Tab.~\ref{tab: det}, FNA surpasses both the manually designed network MobileNetV2 and the NAS-searched network MnasNet-92, where MnasNet~\citep{MnasNet} takes around 3.8K GPU days to search for the backbone network on ImageNet. The specific cost FNA takes on SSDLite is shown in Tab.~\ref{tab: det cost}. It is difficult to train the small network due to the simplification~\citep{liu2019training}. Therefore, experiments on SSDLite need long training schedule and take larger computation cost. The experimental results further demonstrate the efficiency and effectiveness of direct adaptation on the target task with parameter remapping.

\subsection{Effectiveness of Parameter Remapping}
\label{subsec: effect PR}
To evaluate the effectiveness of the parameter remapping paradigm in our method, we attempt to optionally remove the parameter remapping process before the two stages, i.e. architecture adaptation and parameter adaptation. The experiments are conducted with the DeepLabv3~\citep{deeplab-v3} semantic segmentation framework on the Cityscapes dataset~\citep{DBLP:conf/cvpr/CordtsORREBFRS16}. 

\begin{table*}[thbp]
    \vspace{-5pt}
    \centering
    \caption{Effectiveness evaluation of Parameter Remapping. The experiments are conducted with DeepLabv3 on Cityscapes.  Remap: Parameter Remapping. ArchAdapt: Architecture Adaptation. RandInit: Random Initialization. Pretrain: ImageNet Pretrain. ParamAdapt: Parameter Adaptation.}
    \label{tab: effect of PR}
    \begin{threeparttable}
    \resizebox{0.9\textwidth}{!}{
        \begin{tabular}{c l c c c}
        \toprule
        \multirow{1}*{\textbf{Row Num}} & {\textbf{Method}} & \textbf{MAdds(B)} & \textbf{mIOU(\%)}\\
        \midrule
        (1) & Remap $\to$ ArchAdapt $\to$ Remap $\to$ ParamAdapt (FNA) & 24.17 & \textbf{76.6}  \\
        (2) & RandInit $\to$ ArchAdapt $\to$ Remap $\to$ ParamAdapt & 24.29 & 76.0  \\
        (3) & Remap $\to$ ArchAdapt $\to$ RandInit $\to$ ParamAdapt & 24.17 & 73.0  \\
        (4) & RandInit $\to$ ArchAdapt $\to$ RandInit $\to$ ParamAdapt & 24.29 & 72.4  \\
        (5) & Remap $\to$ ArchAdapt $\to$ Retrain $\to$ ParamAdapt & 24.17 & 76.5  \\
        \bottomrule
    \end{tabular}}
    \end{threeparttable}
    \vspace{-5pt}
\end{table*}

In Row (2) we remove the parameter remapping process before architecture adaptation. In other word, the search is performed from scratch without using the pre-trained network. The mIOU in Row (2) drops by 0.6\% compared to FNA in Row (1). Then we remove the parameter remapping before parameter adaptation in Row (3), i.e. training the target architecture from scratch on the target task. The mIOU decreases by 3.6\% compared the result of FNA. When we remove the parameter remapping before both stages in Row (4), it gets the worst performance. In Row (5), we first pre-train the searched architecture on ImageNet and then fine-tune it on the target task. It is worth noting that FNA even achieves a higher mIOU by a narrow margin (0.1\%) than the ImageNet pre-trained one in Row (5). We conjecture that this may benefit from the regularization effect of parameter remapping before the parameter adaptation stage.

All the experiments are conducted using the same searching and training settings for fair comparisons. With parameter remapping applied on both stages, the adaptation achieves the best results in Tab.~\ref{tab: effect of PR}. Especially, the remapping process before parameter adaptation tends to provide greater performance gains than the remapping before architecture adaptation. All the experimental results demonstrate the importance and effectiveness of the proposed parameter remapping scheme.

\begin{table*}[htbp]
    \vspace{-5pt}
    \centering
    \caption{Results of random search experiments with the RetinaNet framework on MS-COCO. DiffSearch: Differentiable NAS. RandSearch: Random Search. The other definitions of abbreviations are the same as Tab.~\ref{tab: effect of PR}.}
    \label{tab: randsearch}
    \begin{threeparttable}
    \resizebox{0.9\textwidth}{!}{
        \begin{tabular}{c l c c c}
        \toprule
        \multirow{1}*{\textbf{Row Num}} & {\textbf{Method}} & \textbf{MAdds(B)} & \textbf{mAP(\%)} \\
        \midrule
        (1) & DetNAS~\citep{DBLP:journals/corr/abs-1903-10979} & 133.26 & 33.3    \\
        (2) & Remap $\to$ DiffSearch $\to$ Remap $\to$ ParamAdapt (FNA) & 133.03 & \textbf{33.9}   \\
        (3) & Remap $\to$ RandSearch $\to$ Remap $\to$ ParamAdapt & 133.11 & 33.5    \\
        (4) & RandInit $\to$ RandSearch $\to$ Remap $\to$ ParamAdapt & 133.08 & 31.5   \\
        (5) & Remap    $\to$ RandSearch $\to$ RandInit $\to$ ParamAdapt & 133.11 & 25.3    \\
        (6) & RandInit  $\to$ RandSearch $\to$ RandInit $\to$ ParamAdapt & 133.08 & 24.9    \\
        \bottomrule
    \end{tabular}}
    \end{threeparttable}
    \vspace{-5pt}
\end{table*}

\subsection{Random Search Experiments}
We carry out the Random Search (RandSearch) experiments with the RetinaNet~\citep{lin2017focal} framework on the MS-COCO~\citep{COCO} dataset. All the results are shown in the Tab.~\ref{tab: randsearch}. We purely replace the original differentiable NAS (DiffSearch) method in FNA with the random search method in Row (3). The random search takes the same computation cost as the search in FNA for fair comparisons. We observe that FNA with RandSearch achieves comparable results with our original method. It further confirms that FNA is a general framework for network adaptation and has great generalization. NAS is only an implementation tool for architecture adaptation. The whole framework of FNA can be treated as a NAS-method agnostic mechanism. It is worth noting that even using random search, our FNA still outperforms DetNAS~\citep{DBLP:journals/corr/abs-1903-10979} with 0.2\% mAP better and 150M MAdds fewer. 

We further conduct similar ablation studies with experiments in Sec.~\ref{subsec: effect PR} about the parameter remapping scheme in Row (4) - (6). All the experiments further support the effectiveness of the parameter remapping scheme.

\subsection{Study on Parameter Remapping}
\label{subsec: studyPR}
We explore more strategies for the Parameter Remapping paradigm. Similar to Sec.~\ref{subsec: effect PR}, all the experiments are conducted with the DeepLabv3~\citep{deeplab-v3} framework on the Cityscapes dataset~\citep{DBLP:conf/cvpr/CordtsORREBFRS16}. We make exploration from the following respects. For simplicity, we denote the weights of the seed network and the new network on the remapping dimension (output/input channel) as $\mW_s = (\mW_s^{(1)} \dots \mW_s^{(p)})$ and $\mW_n = (\mW_n^{(1)} \dots \mW_n^{(q)})$, where $q \le p$.

\begin{figure}[thbp]
    \centering
    \begin{minipage}[b]{0.45\textwidth}
        \centering
        \begin{algorithm}[H]
            \label{algo: mapping}
            \caption{Weights Remapping Function}
            \LinesNumbered
            \KwIn{the seed weights $\mW_s$ and the new \\
            weights $\mW_n$, the reference vector $\vv$}
            // get indices of topk values of the vector\\
            $\va \gets topk$-$indices(\vv, k=q)$\\
            // sort the indices\\
            $sort(\va)$\\
            \For{$i \in 1, 2, \dots, q$}{
                $\mW^{(i)}_n = \mW^{(\va[i])}_s$
            }
            \KwOut{$\mW_n$ with remapped values}
        \end{algorithm}
    \end{minipage}%
    \hspace{15pt}
    \begin{minipage}[htb]{0.4\textwidth}
        \centering
        \includegraphics[width=0.6\textwidth]{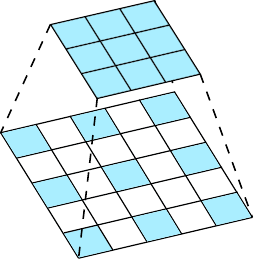}
        \captionof{figure}{Parameter Remapping on the kernel-level with a dilation setting.}
        \label{fig: kernel_dilate}
    \end{minipage}%
    \vspace{-10pt}
\end{figure}

\vspace{-3pt}
\paragraph{Remapping with BN Statistics on Width-level}
We review the formulation of batch normalization~\citep{DBLP:conf/icml/IoffeS15} as follows, 
\begin{equation}
    y_i \gets \gamma \frac{x_i - \mu_{\mathcal{B}}}{\sqrt{\sigma^2_{\mathcal{B}}+\epsilon}} + \beta\text{,}
\end{equation}
where $x_i = (x_i^{(1)} \dots x_i^{(p)})$ denotes the $p$-dimensional input tensor of the $i$th layer, $\gamma \in \R^p$ denotes the learnable parameter which scales the normalized data on the channel dimension. We compute the absolute values of $\gamma$ as $|\gamma| = (|\gamma^{(1)}| \dots |\gamma^{(p)}|)$. When remapping the parameters on the width-level, we sort the values of $|\gamma|$ and map the parameters with the sorted top-$q$ indices. More specifically, we define a weights remapping function in Algo.~\ref{algo: mapping}, where the reference vector $\vv$ is $|\gamma|$.

\vspace{-5pt}
\paragraph{Remapping with Weight Importance on Width-level}
We attempt to utilize a canonical form of convolution weights to measure the importance of parameters. Then we remap the seed network parameters with great importance to the new network. The remapping operation is conducted based on Algo.~\ref{algo: mapping} as well. We experiment with two canonical forms of weights to compute the reference vector, the standard deviation of $\mW_s$ as $(std(\mW_s^{(1)}) \dots std(\mW_s^{(p)}))$ and the $\normlone$ norm of $\mW_s$ as $(|| \mW_s^{(1)} ||_1 \dots || \mW_s^{(p)} ||_1)$.

\begin{table*}[htbp]
    \vspace{-10pt}
    \centering
    \caption{Study the methods of Parameter Remapping.}
    \label{tab: methods of PR}
    \begin{threeparttable}
    \resizebox{0.75\textwidth}{!}{
        \begin{tabular}{l c c c c c}
        \toprule
        \textbf{Method} & Width-BN & Width-Std & Width-L1 & Kernel-Dilate & FNA \\
        \midrule
        \textbf{mIOU(\%)} & 75.8 & 75.8 & 75.3 & 75.6 & \textbf{76.6} \\
        \bottomrule
    \end{tabular}}
    \end{threeparttable}
    \vspace{-5pt}
\end{table*}

\vspace{-5pt}
\paragraph{Remapping with Dilation on Kernel-level}
We experiment with another strategy of parameter remapping on the kernel-level. Different from the function-preserving method defined in Sec.~\ref{subsec: params_remap}, we remap the parameters with a dilation manner as shown in Fig.~\ref{fig: kernel_dilate}. The values in the convolution kernel without remapping are all assigned as $0$. It is formulated as
\begin{equation}
    \mW^{k \times k}_{:,:,h,w} = \begin{cases}
        \mW^{3 \times 3}_{:,:,h,w} & \text{if } h, w = 1 + i \cdot \frac{k-1}{2} \text{ and } i=0,1,2\\
        0 & \text{otherwise}
    \end{cases}\text{,}
    \label{eq: kernel_dilate}
\end{equation}
where $\mW^{k \times k}$ and $\mW^{3 \times 3}$ denote the weights of the new network and the seed network respectively, $h, w$ denote the spatial indices.

Tab.~\ref{tab: methods of PR} shows the experimental results. The network adaptation with the parameter remapping paradigm define in FNA achieves the best results. Furthermore, the remapping operation of FNA is easier to implement compared to the several aforementioned ones. However, we explore limited number of methods to implement the parameter remapping paradigm. How to conduct the remapping strategy more efficiently remains a significative future work.

% --------------------------------Section Partition------------------------------- %
\section{Conclusion}
In this paper, we propose a fast neural network adaptation method (FNA) with a parameter remapping paradigm and the architecture search method. We adapt the manually designed network MobileNetV2 to semantic segmentation and detection tasks on both architecture- and parameter- level. The parameter remapping strategy takes full advantages of the seed network parameters, which greatly accelerates both the architecture search and parameter fine-tuning process. With our FNA method, researchers and engineers could fast adapt more manually designed networks to various frameworks on different tasks. As there are lots of ImageNet pre-trained models available in the community, we could conduct adaptation with low cost and do more applications, e.g., face recognition, pose estimation, depth estimation, etc. We leave more efficient remapping strategies and more applications for future work.

\section*{Acknowledgement}
This work was supported by National Natural Science Foundation of China (NSFC) (No. 61876212, No. 61733007 and No. 61572207), National Key R\&D Program of China (No. 2018YFB1402600) and HUST-Horizon Computer Vision Research Center. We thank Liangchen Song, Yingqing Rao and Jiapei Feng for the discussion and assistance.

\bibliography{iclr2020_conference}
\bibliographystyle{iclr2020_conference}

\appendix
\section{Appendix}
\subsection{Implementation Details on Semantic Segmentation}
For architecture adaptation, the image is first resized to $512 \times 1024$ and $321 \times 321$ patches are randomly cropped as the input data. The output feature maps are down-sampled by the factor of $16$. Depthwise separable convolutions are used in the ASPP module~\citep{DBLP:journals/pami/ChenPKMY18, deeplab-v3}. The stages where the expansion ratio of MBConv is 6 in the original MobileNetV2 are searched and adjusted. We set the maximum numbers of layers in each searched stage of the super network as $\{4, 4, 6, 6, 4, 1\}$. We set a warm-up stage in the first $5$ epochs to linearly increase the learning rate from $1 \times 10^{-4}$ to $0.02$. Then, the learning rate decays to $1 \times 10^{-3}$ with the cosine annealing schedule~\citep{DBLP:conf/iclr/LoshchilovH17}. The batch size is set as $16$. We use the SGD optimizer with $0.9$ momentum and $5 \times 10^{-4}$ weight decay for operation weights and the Adam optimizer~\citep{DBLP:journals/corr/KingmaB14} with $4 \times 10^{-5}$ weight decay and a fixed learning rate $1 \times 10^{-3}$ for architecture parameters.

For parameter adaptation, the training data is cropped as a $769 \times 769$ patch from the rescaled image. The scale is randomly selected from $[0.75, 1.0, 1.25, 1.5, 1.75, 2.0]$. The random left-right flipping is used. We update the statistics of the batch normalization (BN)~\citep{DBLP:conf/icml/IoffeS15} for $2000$ iterations before the fine-tuning process. We use the same SGD optimizer as the search process. The learning rate linearly increases from $1 \times 10 ^{-4}$ to $0.01$ and then decays to $0$ with the polynomial schedule. The batch size is set as $16$.

\subsection{Implementation Details on Object Detection}
\paragraph{RetinaNet} 
We describe the details in the search process of architecture adaptation as follows. The depth settings in each searched stage are set as $\{4, 4, 4, 4, 4, 1\}$. For the input image size, the short side is resized to 800 while the maximum long side is set as 1333. For operation weights, we use the SGD optimizer with $1 \times 10^{-4}$ weight decay and $0.9$ momentum. We set a warm-up stage in the first $500$ iterations to linearly increase the learning rate from $0$ to $0.02$. Then we decay the learning rate by a factor of $0.1$ at the 8th and 11th epoch. For the architecture parameters, we use the Adam optimizer~\citep{DBLP:journals/corr/KingmaB14} with $1 \times 10^{-3}$ weight decay and a fixed learning rate $3 \times 10^{-4}$. For the multi-objective loss function, we set $\lambda$ as $0.02$ and $\tau$ as $10$. We begin optimizing the architecture parameters after 8 epochs. We remove the random flipping operation on input images in the search process. All the other training settings are the same as MMDetection~\citep{chen2019mmdetection} implementation. 

For fine-tuning of the parameter adaptation, we use the SGD optimizer with $5 \times 10^{-5}$ weight decay and 0.9 momentum. A similar warm-up procedure is set in the first $500$ iterations to increase the learning rate from $0$ to $0.05$. Then we decay the learning rate by $0.1$ at the 8th and 11th epoch. The whole architecture search process takes $14$ epochs, $18$ hours on 8 TITAN-Xp GPUs with the batch size of 8 and the whole parameter fine-tuning takes 12 epochs, $10$ hours on 8 TITAN-Xp GPUs with 32 batch size.

\paragraph{SSDLite}
We resize the input images to $320 \times 320$. For operation weights in the search process, we use the standard RMSProp optimizer with $4 \times 10^{-5}$ weight decay. The warm-up stage in the first $500$ iterations increases learning rate from $0$ to $0.03$. Then we decay the learning rate by $0.1$ at 16 and 22 epochs. The architecture optimization starts at 12 epochs. We set $\lambda$ as $0.2$ and $\tau$ as $10$ for the loss function. The other search settings are the same as the RetinaNet experiment. 

For parameter adaptation, the initial learning rate is $0.2$ and decays at 36, 50 and 56 epochs. The training settings follow the SSD~\citep{liu2016ssd} implementation in MMDetection~\citep{chen2019mmdetection}. The search process takes 24 epochs in total, $20$ hours on 8 TITAN-Xp GPUs with 64 batch size. The parameter adaptation takes 60 epochs, $46$ hours on 8 TITAN-Xp GPUs with 512 batch size.

\begin{figure*}[htbp]
    \centering
    \subfigure[FNA on DeepLabv3]{
    \begin{minipage}[b]{1\linewidth}
        \includegraphics[width=1\linewidth]{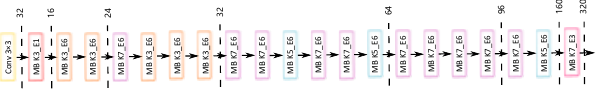}
    \end{minipage}}
    \subfigure[FNA on RetinaNet]{
    \begin{minipage}[b]{1\linewidth}
        \includegraphics[width=1\linewidth]{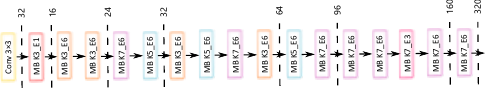}
    \end{minipage}}
    \subfigure[FNA on SSDLite]{
    \begin{minipage}[b]{1\linewidth}
        \includegraphics[width=1\linewidth]{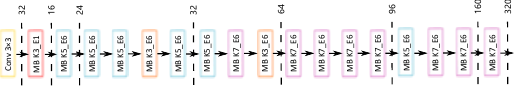}
    \end{minipage}}
    \caption{Visualization of our searched architectures on different frameworks. MB: inverted residual block proposed in MobilenetV2~\citep{sandler2018mobilenetv2}. Kx\_Ey: the kernel size of the depthwise convolution is x and the expansion ratio is y.}
\end{figure*}

\begin{figure*}[hbp]
    \centering
    \subfigure[Input]{
    \begin{minipage}[b]{0.31\linewidth}
        \includegraphics[width=1\linewidth]{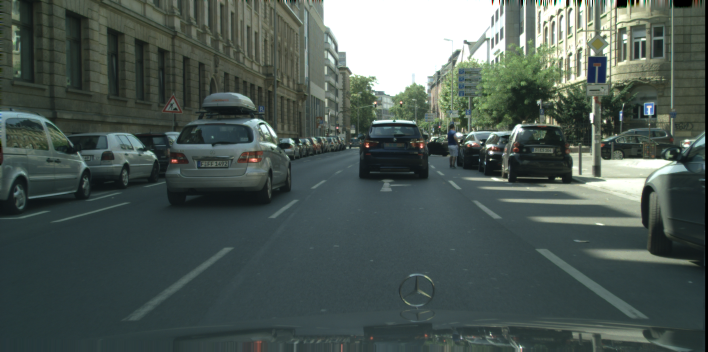}\vspace{4pt}
        \includegraphics[width=1\linewidth]{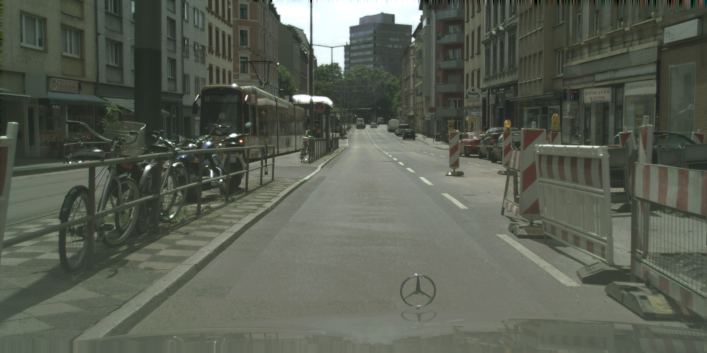}\vspace{4pt}
        \includegraphics[width=1\linewidth]{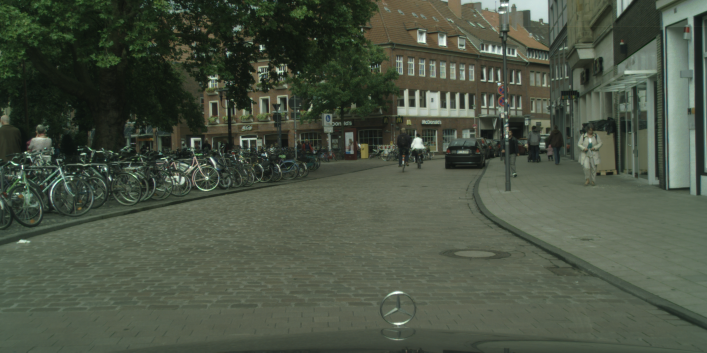}\vspace{4pt}
    \end{minipage}}
    \subfigure[GT]{
    \begin{minipage}[b]{0.31\linewidth}
        \includegraphics[width=1\linewidth]{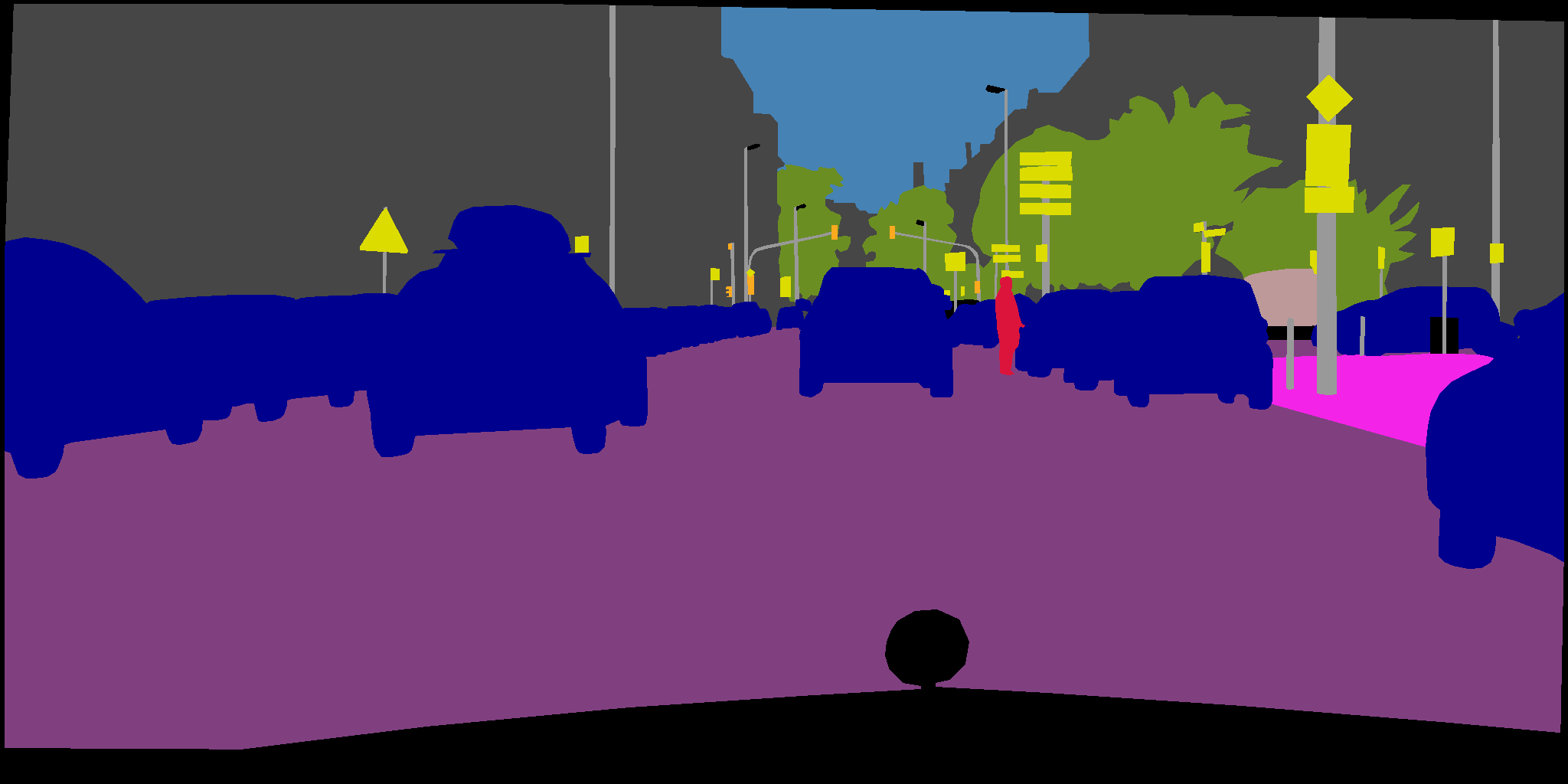}\vspace{4pt}
        \includegraphics[width=1\linewidth]{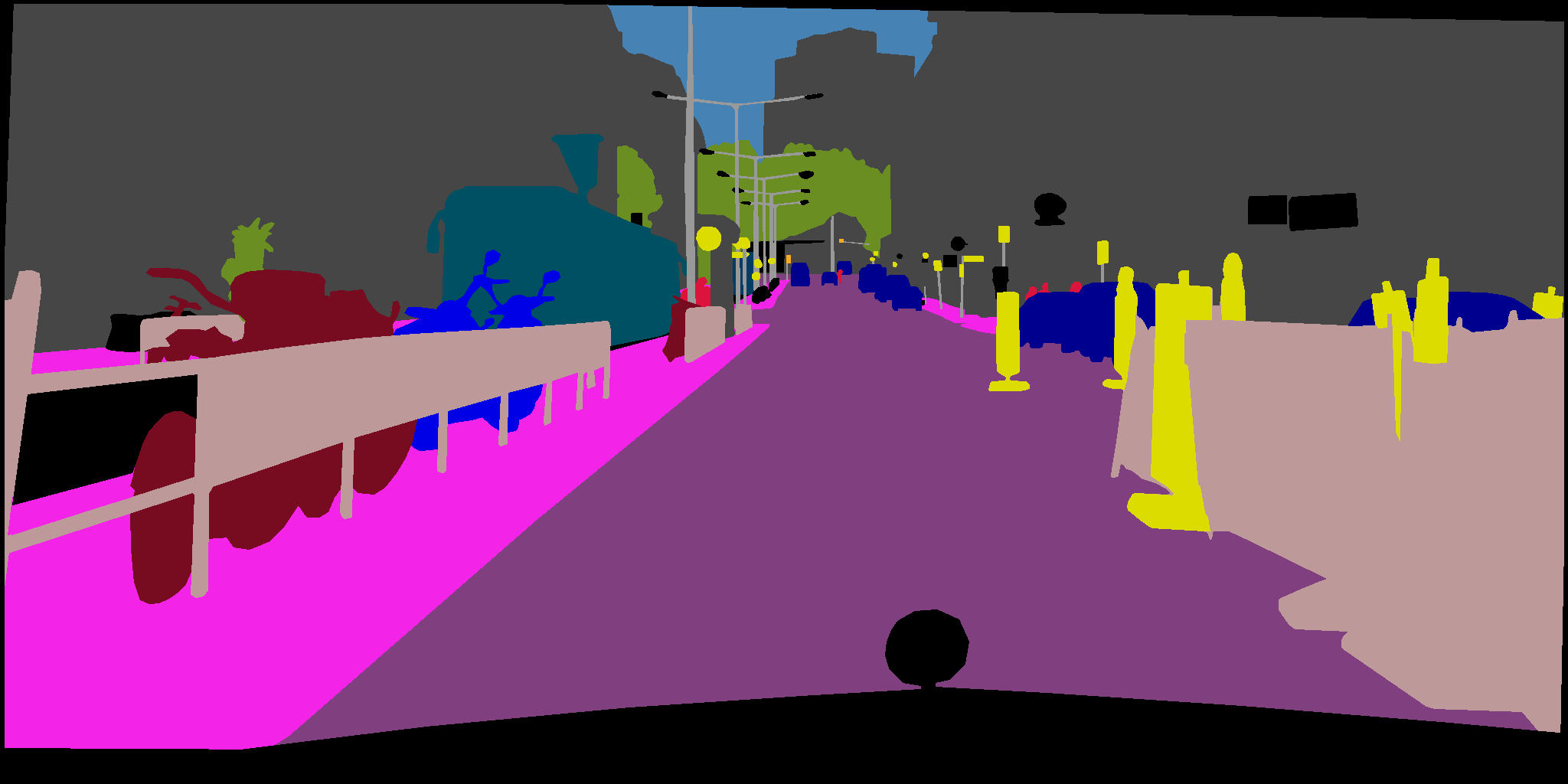}\vspace{4pt}
        \includegraphics[width=1\linewidth]{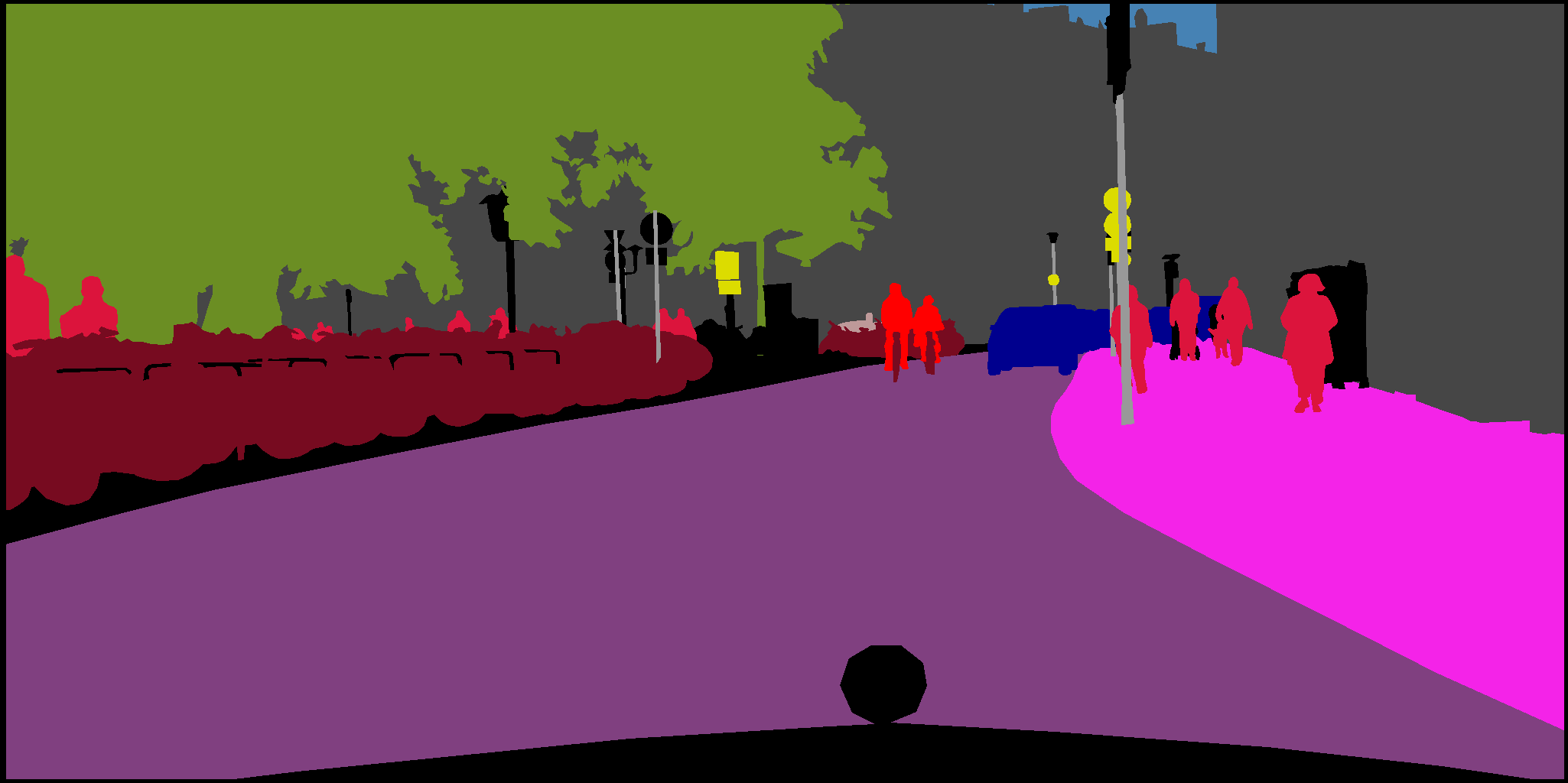}\vspace{4pt}
    \end{minipage}}
    \subfigure[Ours]{
    \begin{minipage}[b]{0.31\linewidth}
        \includegraphics[width=1\linewidth]{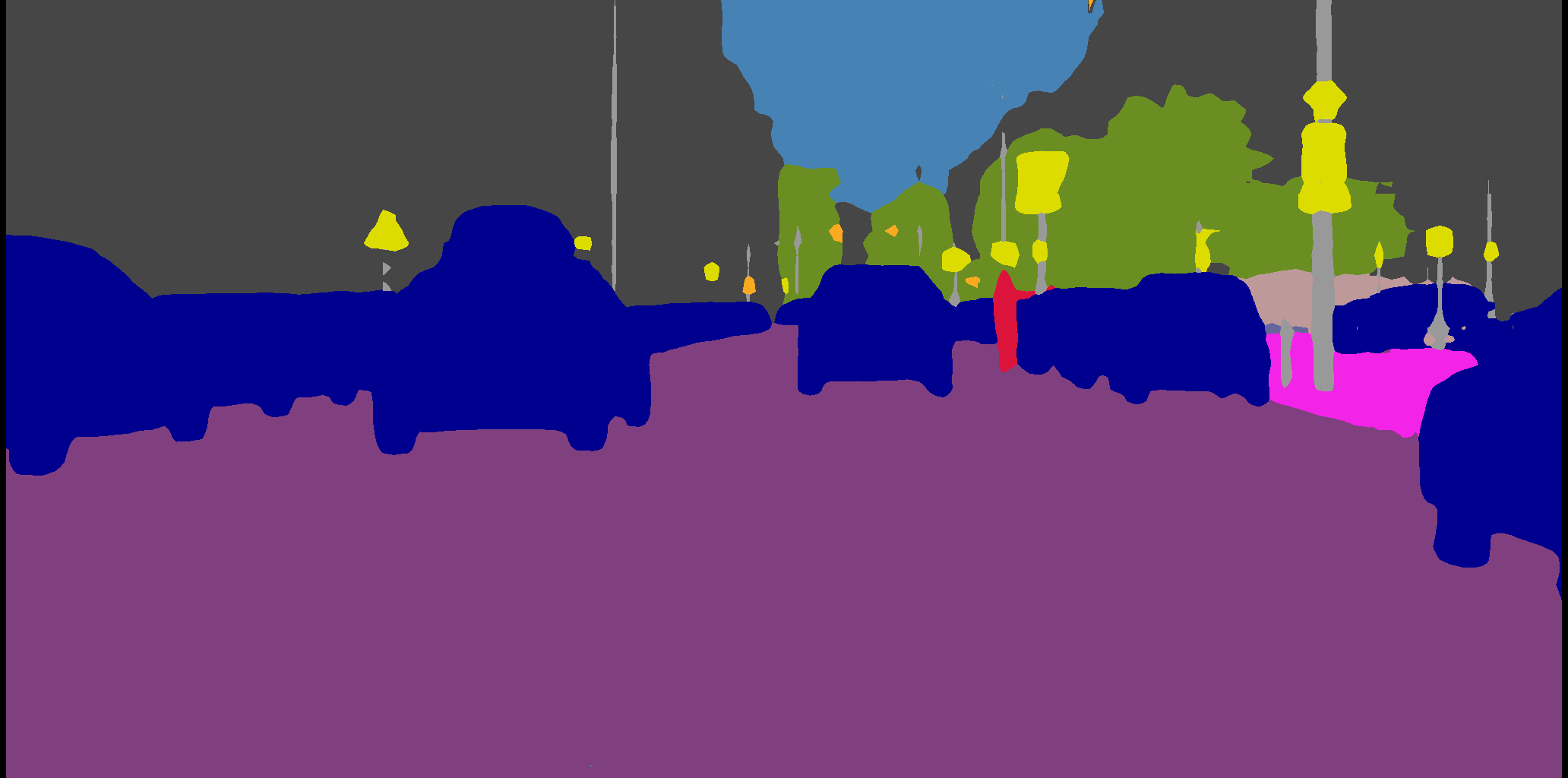}\vspace{4pt}
        \includegraphics[width=1\linewidth]{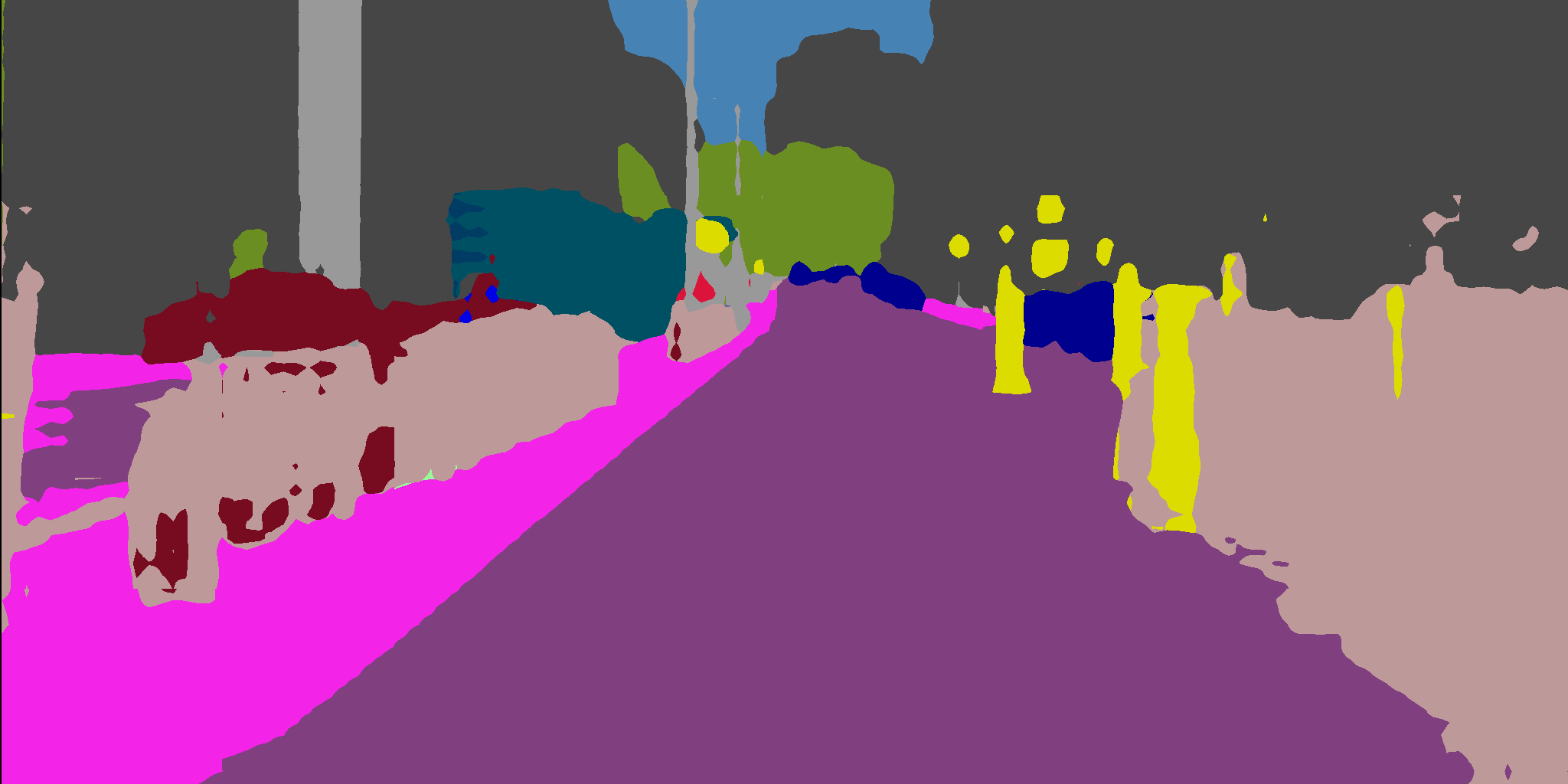}\vspace{4pt}
        \includegraphics[width=1\linewidth]{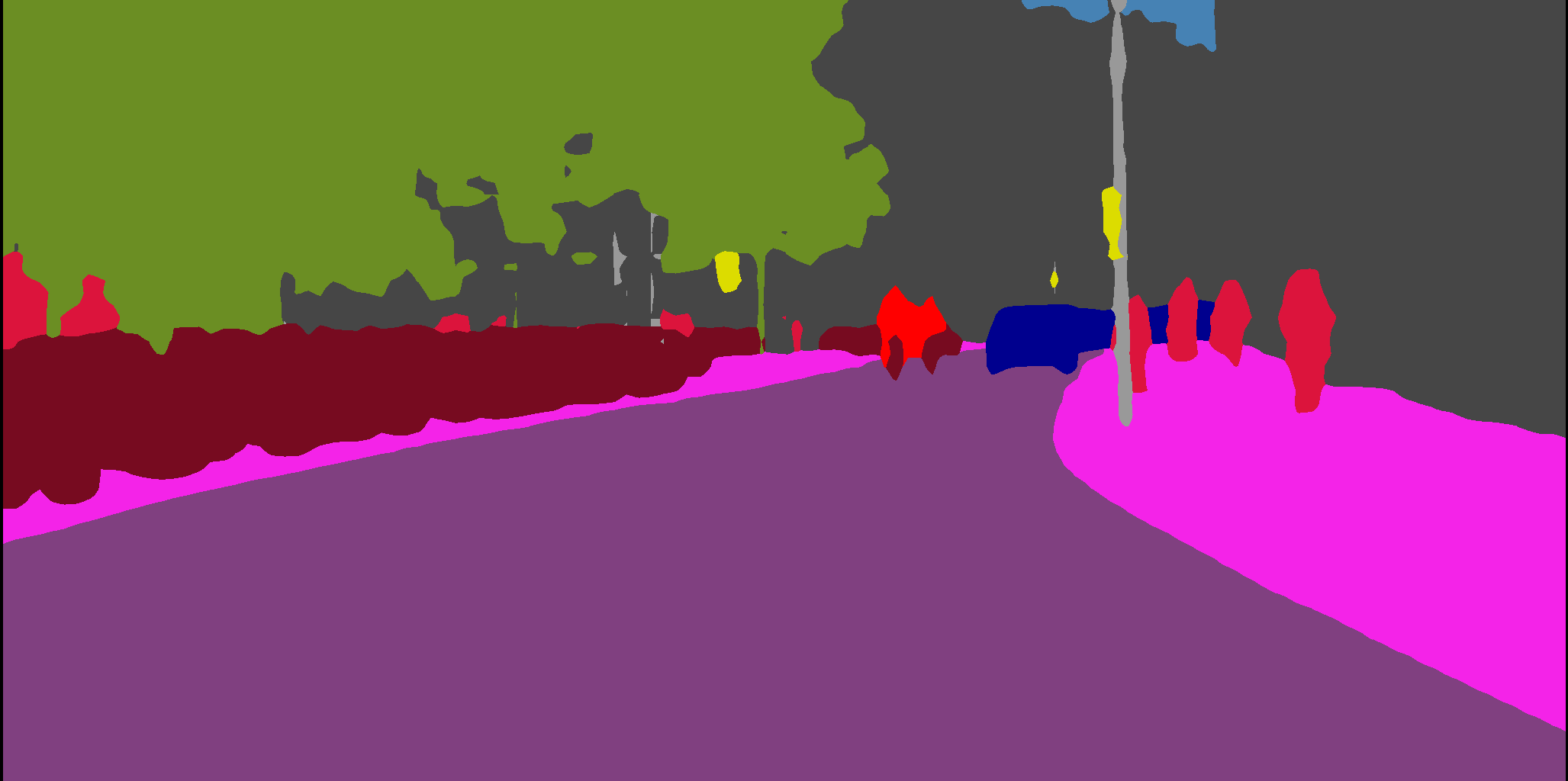}\vspace{4pt}
    \end{minipage}}
    \caption{Visualization of semantic segmentation results on the Cityscapes validation dataset.}
\end{figure*}

\begin{figure*}[hbp]
    \centering
    \subfigure{
    \begin{minipage}[b]{0.31\linewidth}
        \includegraphics[width=1\linewidth]{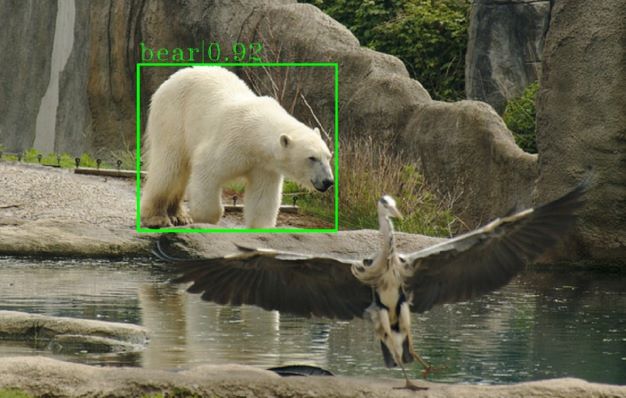}\vspace{4pt}
        \includegraphics[width=1\linewidth]{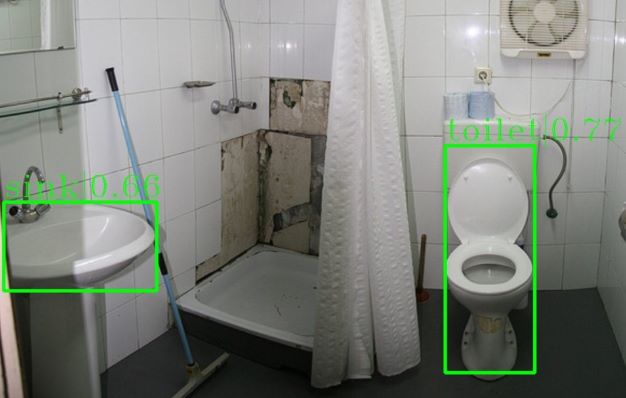}\vspace{4pt}
        \includegraphics[width=1\linewidth]{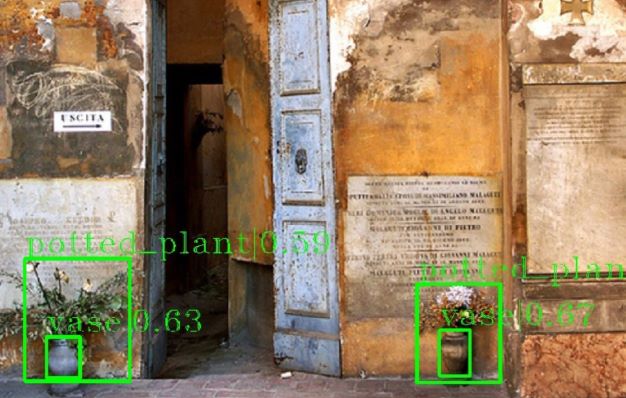}
    \end{minipage}}
    \subfigure{
    \begin{minipage}[b]{0.31\linewidth}
        \includegraphics[width=1\linewidth]{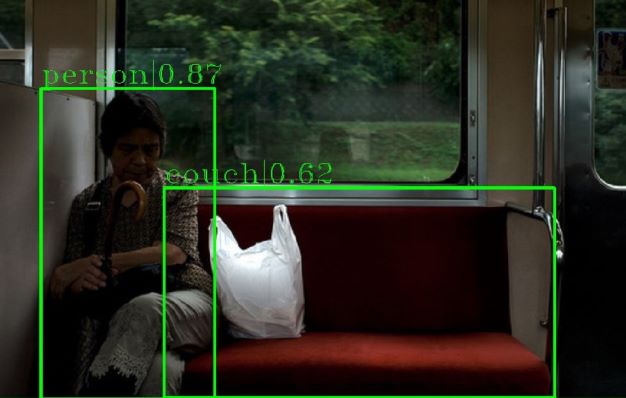}\vspace{4pt}
        \includegraphics[width=1\linewidth]{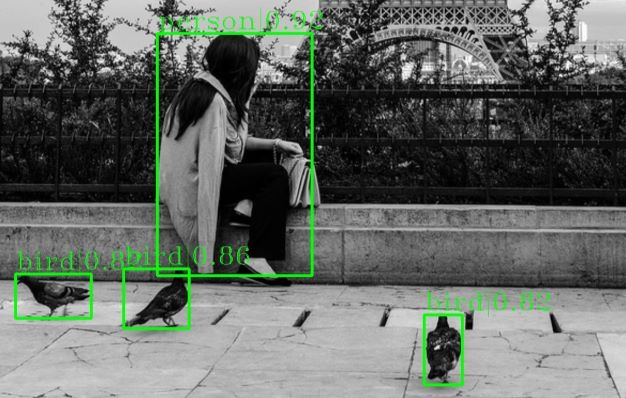}\vspace{4pt}
        \includegraphics[width=1\linewidth]{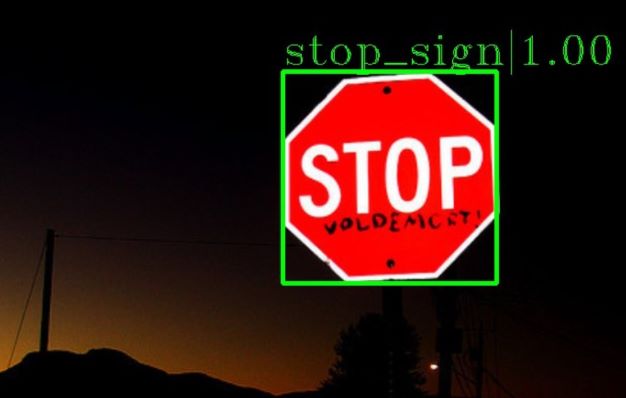}
    \end{minipage}}
    \subfigure{
    \begin{minipage}[b]{0.31\linewidth}
        \includegraphics[width=1\linewidth]{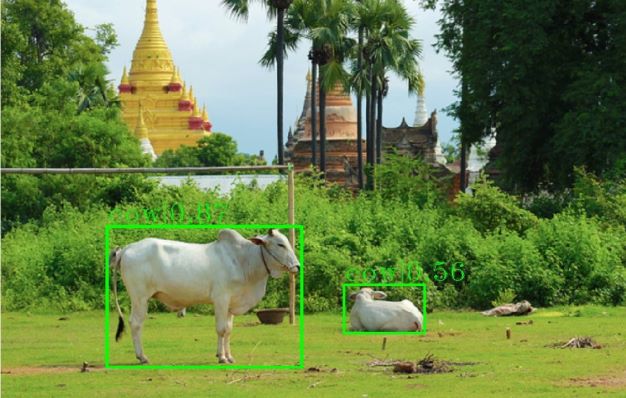}\vspace{4pt}
        \includegraphics[width=1\linewidth]{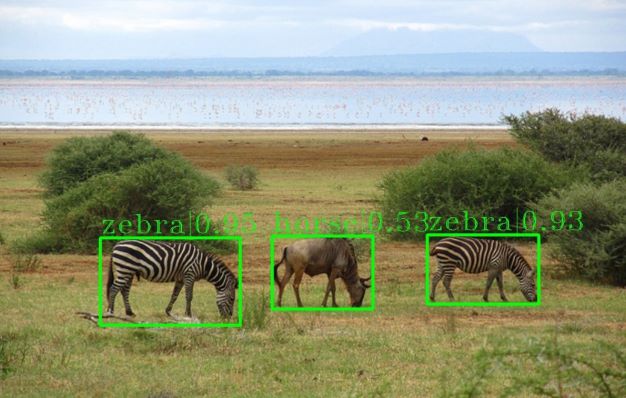}\vspace{4pt}
        \includegraphics[width=1\linewidth]{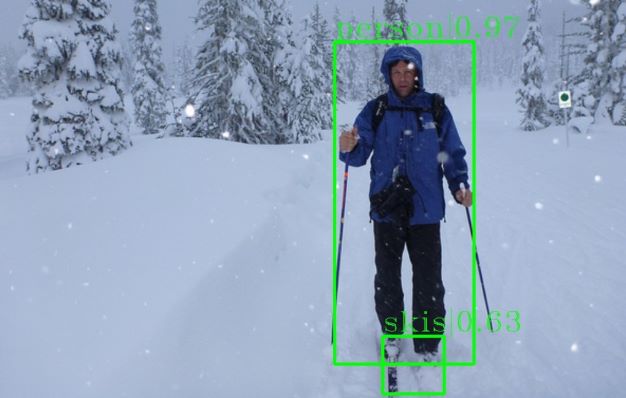}
    \end{minipage}}
    \caption{Visualization of object detection results on the MS-COCO validation dataset.}
\end{figure*}

\end{document}